\documentclass[twoside]{article}

\usepackage[accepted]{collapsing_FLAME/arxiv/aistats2019}
\usepackage[round]{natbib}
\newcommand\blfootnote[1]{%
  \begingroup
  \renewcommand\thefootnote{}\footnote{#1}%
  \addtocounter{footnote}{-1}%
  \endgroup
}

\newcommand{\eat}[1]{{}}

\usepackage[utf8]{inputenc} 
\usepackage[T1]{fontenc}    
\usepackage{hyperref}       
\usepackage{url}            
\usepackage{booktabs}       
\usepackage{amsfonts}       
\usepackage{nicefrac}       
\usepackage{microtype}      
\usepackage{amsmath}
\usepackage{multicol}
\usepackage{bbm}
\usepackage{enumitem}
\usepackage{xcolor}
\usepackage[ruled,vlined]{algorithm2e}
\usepackage{wrapfig}
\usepackage{graphicx}
\usepackage{caption}
\usepackage{subfloat}
\usepackage{amsthm}
\usepackage{float}
\usepackage{subcaption}
\usepackage{pdfpages}
\usepackage{footmisc}
\usepackage{balance}
\usepackage{MnSymbol}
\usepackage{bm}
\usepackage{bbm}

\newcommand{\iter}{{h}} 
\newcommand{\btheta}{\bm{\theta}}

\newcommand{\x}{\mathbf{x}}
\newcommand{\vv}{\boldsymbol{\bm{\theta}}}

\newcommand{\MG}{{\cal{MG}}}

\definecolor{LightCyan}{rgb}{0.88,1,1}
\definecolor{Gray}{gray}{0.9}

\newcommand{\proj}[1]{{\Pi}}
\newcommand{\sel}[1]{{\sigma}}

\newcommand{\cut}[1]{}
\newcommand{\cutfull}[1]{}

\newcommand{\commentresolved}[1]{}

\usepackage{aliascnt}  		

\newtheorem{theorem}{Theorem}[section]          	
 \newaliascnt{lemma}{theorem}				
 \aliascntresetthe{lemma}  					
\newaliascnt{proposition}{theorem}			
\newtheorem{proposition}[proposition]{Proposition}  
\aliascntresetthe{proposition}  				

\cut{

\providecommand{\algorithmautorefname}{Algorithm}

}

\newcommand{\BasicExactMatch}{{\tt GroupedMR}}
\newcommand{\collapsingflame}{{\tt DAME}}
\newcommand{\GenerateNewActiveSets}{{\tt GenerateNewActiveSets}}

\newcommand{\PE}{{\tt PE}}

\newcommand{\w}{\mathbf{w}}

\begin{document}

%

%

\twocolumn[


\aistatstitle{Interpretable Almost Matching Exactly for Causal Inference}


%
\aistatsauthor{
  Yameng Liu, Awa Dieng,
  Sudeepa Roy, Cynthia Rudin, Alexander Volfovsky${^*}$
}
\aistatsaddress{
  Duke University\\ 
  Durham, NC 27708\\ 
  \texttt{\texttt{$\{$awa.dieng, alexander.volfovsky$\}$@duke.edu, $\{$ymliu, sudeepa, cynthia$\}$@cs.duke.edu}}
}
]

\begin{abstract}
Matching methods are heavily used in the social and health sciences due to their interpretability. We aim to create the highest possible quality of treatment-control matches for categorical data in the potential outcomes framework.  
The method proposed in this work aims to match units on a weighted Hamming distance, taking into account the relative importance of the covariates; the algorithm aims to match units on as many relevant variables as possible. To do this, the algorithm creates a hierarchy of covariate combinations on which to match (similar to downward closure), in the process solving an optimization problem for each unit in order to construct the optimal matches. The algorithm uses a single dynamic program to solve all of the units' optimization problems simultaneously. Notable advantages of our method over existing matching procedures are its high-quality interpretable matches, versatility in handling different data distributions that may have irrelevant variables, and ability to handle missing data by matching on as many available covariates as possible.
\end{abstract}
\section{INTRODUCTION}\label{sec:intro}
\blfootnote{$^*$ Equal contribution from all authors.}In observational causal inference where the scientist does not control the randomization of individuals into treatment, an ideal approach
matches each treatment unit to a control unit with identical covariates. However, in high dimensions, few such ``identical twins'' exist, since it becomes unlikely that any two units have identical covariates in high dimensions. In that case, how might we construct a match assignment that would lead to accurate estimates of conditional average treatment effects (CATEs)?

For categorical variables, we might choose a Hamming distance to measure similarity between covariates. Then, the goal is to find control units that are similar to the treatment units on as many covariates as possible. 
However, the fact that not all covariates are equally important has serious implications for CATE estimation. Matching methods generally suffer in the presence of many irrelevant covariates (covariates that are not related to either treatment or outcome): the irrelevant variables would dominate the Hamming distance calculation, so that the treatment units would mainly be matched to the control units on the irrelevant variables. This means that matching methods do not always pass an important sanity check in that irrelevant variables should be irrelevant. 
To handle this issue with irrelevant covariates, in this work we choose to match units based on a \textit{weighted} Hamming distance, where the weights can be learned from machine learning on a hold-out training set. These weights act like variable importance measures for defining the Hamming distance.

The choice to optimize matches using Hamming distance leads to a serious computational challenge: how does one compute optimal matches on Hamming distance? In this work, we define a matched group for a given unit as the solution to a constrained discrete optimization problem, which is to find the weighted Hamming distance of each treatment unit to the nearest control unit (and vice versa).  
There is one such optimization problem for each unit, and we solve all of these optimization problems efficiently with a single dynamic program. Our dynamic programming algorithm has the same basic monotonicity property (downwards closure) as that of the apriori algorithm \citep{Agrawal1994} used in data mining for finding frequent itemsets. However, frequency of itemsets is irrelevant here, instead the goal is to find a largest (weighted) set of covariates that both a treatment and control unit have in common. The algorithm, Dynamic Almost Matching Exactly -- \collapsingflame\ -- is efficient, owing to the use of 
bit-vector computations to match units in groups, and does not require an integer programming solver. 

A more general version of our formulation (Full Almost Matching Exactly) adaptively chooses the features for matching in a data-driven way. Instead of using a fixed weighted Hamming distance, it uses the hold-out training set to determine how useful a \textit{set} of variables is for prediction out of sample. For each treatment unit, it finds a set of variables that (i) allows a match to at least one control unit; (ii) together have the best out-of-sample prediction ability among all subsets of variables for which a match can be created (to at least one control unit). Again, even though for each unit we are searching for the best subset of variables, we can solve all of these optimization problems at once with our single dynamic program.

\section{RELATED WORK}



As mentioned earlier, exact matching is not possible in high dimensions, as ``identical twins'' in treatment and control samples are not likely to exist. Early on, this led to techniques that reduce dimension using propensity score matching \citep{rubin1973use,rubin1973matching, rubin1976multivariate, cochran1973controlling}, which extend to penalized regression approaches \citep{schneeweiss2009high, rassen2012using, belloni2014inference, farrell2015robust}. Propensity score matching methods project the entire dataset to one dimension and thus cannot be used for estimating CATE (conditional average treatment effect), since units within the matched groups often differ on important covariates. In ``optimal matching,'' \citep{rosenbaum2016imposing}, an optimization problem is formed to choose matches according to a pre-defined distance measure, though as discussed above, this distance measure can be dominated by irrelevant covariates, leading to poor matched groups and biased estimates. Coarsened exact matching \citep{iacus2011causal,iacus2011multivariate} has the same problem, since again, the distance metric is pre-defined, rather than learned. Recent integer-programming-based methods considers extreme matches for all possible reasonable distance metrics, but this is computationally expensive and relies on manual effort to create the ranges \citep{MorucciNoRu18,NoorRu15}; in contrast we use machine learning to create a single good match assignment.
 
 
In the framework of \textit{almost-exact matching} \citep{genericFLAME2017}, each matched group contains units that are close on covariates that are important for predicting outcomes. 
For example, Coarsened Exact Matching \citep{iacus2011causal,iacus2011multivariate} is almost-exact if one were to use an oracle (should one ever become available) that bins covariates according to importance for estimating causal effects.
\collapsingflame's predecessor, the FLAME algorithm \citep{genericFLAME2017} is an almost-exact matching method that adapts the distance metric to the data using machine learning. It starts by matching ``identical twins,'' and proceeds by eliminating less important covariates one by one, attempting to match individuals on the largest set of covariates that produce valid matched groups. 
FLAME can handle huge datasets, even datasets that are too large to fit in memory, and scales well with the number of covariates, but removing covariates in exactly one order (rather than all possible orders as in \collapsingflame) means that many high-quality matches will be missed. 

\collapsingflame\ tends to match on more covariates than FLAME; the distances between matched units are smaller in \collapsingflame\ than in FLAME, thus its matches are distinctly higher quality. This has implications for missing data, where \collapsingflame\ can find matched groups that FLAME cannot.

\section{ALMOST MATCHING EXACTLY \\(AME) FRAMEWORK}
\label{sec:AEM}

Consider a dataframe $D = [X,Y,T]$ where $X\in \{0,1,\dots,k\}^{n\times p}$, $Y\in \mathbb{R}^n$, $T\in\{0,1\}^n$ respectively denote the categorical covariates for all units, the outcome vector and the treatment indicator ($1$ for treated, $0$ for control). The $j$-th covariate $X$ of unit $i$ is denoted $x_{ij}\in \{0,1,\dots,k\}$. Notation $\x_i\in \{0,1,\dots,k\}^p$ indicates covariates for the $i$th unit, and $T_i\in\{0,1\}$ is an indicator for whether or not unit $i$ is treated.

Throughout we make SUTVA and ignorability assumptions \citep{Rubin1980stuva}. 
The goal is to match treatment and control units on as many relevant covariates as possible. Relevance of covariate $j$ is denoted by $w_j\geq 0$ and it is determined using a hold-out training set. $w_j$'s can either be fixed beforehand or adjusted dynamically inside the algorithm (see Full-AME in Section \ref{sec:adaptive}). 

For now, assuming that we have a fixed nonnegative weight $w_j$ for each covariate $j$, we would like to find a match for each treatment unit $t$ that \textit{matches at least one control unit on as many relevant covariates as possible}. Thus we consider the following problem: 

\noindent\textbf{Almost Matching Exactly with Fixed Weights (AME):}
\textit{For each treatment unit} $t$, 
\begin{eqnarray*}
\lefteqn{\vv^{t*} \in \mathrm{argmax}_{\vv\in\{0,1\}^p} \vv^T\mathbf{w} \textit{ such that }} \\
&&\exists \;\ell \;\textit{ with }T_{\ell}=0 \textit{ and } \x_{\ell}\circ \vv = \x_t\circ\vv,  
\end{eqnarray*}
where $\circ$ denotes Hadamard product. The solution to the AME problem is an indicator of the optimal set of covariates for the matched group of treatment unit $t$. The constraint says that the optimal matched group contains at least one control unit.  
When the solution of the AME problem is the same for multiple treatment units, they form a single matched group.
For treatment unit $t$, the \textbf{main matched group} for $t$ contains all units $\ell$ so that $\x_t\circ \vv^{t*} = \x_{\ell}\circ \vv^{t*}$. If any unit $\ell$ (either control or treatment) within $t$'s main matched group has its own different main matched group, then $t$'s matched group is an \textbf{auxiliary matched group} for $\ell$. In this case, $\ell$ could have been matched to other units on more covariates than it was matched to $t$. Estimation of CATE for a unit should always be done on the main matched group for that unit.

The formulation of the AME and main matched group is symmetric for control units. There are two straightforward (but inefficient) approaches to solving the AME problem for all units.

\noindent \textbf{AME Solution 1 (quadratic in $n$, linear in $p$):} Brute force pairwise comparison of treatment points to control points. (Detailed in the appendix.)

\noindent \textbf{AME Solution 2 (order $n \log n$, exponential in $p$):} Brute force iteration over all $2^p$ subsets of the $p$ covariates. (Detailed in the appendix.)

If $n$ is in the millions, the first solution, or any simple variation of it, is practically infeasible. A straightforward implementation of the second solution is also inefficient. 
However, a monotonicity property (downward closure) 
allows us to prune the search space so that the second solution can be modified to be completely practical.  
The \collapsingflame\ algorithm 
does not enumerate all $\vv$'s, monotonicity reduces the number of $\vv$'s it considers.
\begin{proposition}\label{prop:downward-closure}
\textrm{\rm (Monotonicity of $\vv^*$ in AME solutions)} 
Fix treatment unit $t$. Consider feasible $\vv$, meaning $\exists \;\ell \;\textit{ with }T_{\ell}=0 \textit{ and } \x_\ell\circ \vv = \x_t\circ\vv$. Then, 
\begin{itemize}[leftmargin=*]
\itemsep0em
\item Any feasible $\vv'$ such that $\vv'<\vv$ elementwise will have $\vv'^T\mathbf{w} \leq  \vv^T\mathbf{w}$.
\item Consequently, consider feasible vectors $\vv$ and $\vv'$. Define $\tilde{\vv}$ as the elementwise $\min(\vv,\vv')$. Then $\tilde{\vv}^T\mathbf{w}<\vv^T \mathbf{w}$, and $\tilde{\vv}^T\mathbf{w}<\vv'^T\mathbf{w}$.
\end{itemize} 
\end{proposition}
These follow from the fact that the elements of $\vv$ are binary and the elements of $\mathbf{w}$ are non-negative. The first property means that if we have found a feasible $\vv$, we do not need to consider any $\vv'$ with fewer 1's as a possible solution of the AME for unit $t$. Thus, the \collapsingflame\ algorithm starts from $\vv$ being all 1's (consider all covariates). It systematically drops one element of $\vv$ to zero at a time, then two, then three, ordered according to values of $\vv^T\mathbf{w}$. 
The second property implies that we must evaluate both $\vv$ and $\vv'$ as possible AME solutions before evaluating $\tilde{\vv}$. Conversely, a new subset of variables defined by $\tilde{\vv}$ cannot be considered unless all of its supersets have been considered. These two properties form the basis of the \collapsingflame\ algorithm. 

The algorithm must be stopped early to avoid creating low quality matches. A useful stopping criterion is if the weighted sum of covariates $\btheta^T\mathbf{w}$ used for matching becomes too low (perhaps lower than a prespecified percentage of the total sum of weights $\|\mathbf{w}\|_1$).

\emph{Note that matching does not produce estimates, it produces a partition of the covariate space, based on which we can estimate CATEs}.
Within each main matched group, we use the difference of the average outcome of the treated units and the average outcome of the control units as an estimate of the CATE value, given the covariate values for that group. Smoothing the CATE estimates could be useful after matching.

\cut{
We now have ingredients for \collapsingflame: the dynamic programming mechanism over subsets of variables and fast implementations of \BasicExactMatch\ using database queries and bit vectors. 
}
\section{DYNAMIC ALMOST MATCHING EXACTLY (DAME)} 
\label{sec:Alg}


We call a {\em covariate-set} any set of covariates. We denote by $\mathcal{J}$ the original set of all covariates from the input dataset, where $p = |\mathcal{J}|$. When we \textit{drop} a set of covariates $s$, it means we will match on $\mathcal{J} \setminus s$.
For any covariate-set $s$, we associate an \emph{indicator-vector} $\vv_s \in\{0,1\}^p$ defined as follows: 
\begin{equation}\label{equn:vector-s}
\vv_{s, j} = \mathbbm{1}_{\{j \notin s\}}~~~~\forall~j \in \{1,..,p\}
\end{equation}
that is, the value is 1 if the covariate is \emph{not in} $s$ implying that it is being used for matching. 


\begin{algorithm}[t] 
  \caption{The \collapsingflame\ algorithm}
  \SetKwInOut{Input}{Input}
  \SetKwInOut{Output}{Output}  

    \Input{ Data $D$, 
    pre-computed weight vector $w$ for all covariates (from machine learning)}
    \Output{$\{D_{(\iter)}^m, \mathcal{MG}_{(\iter)}\}_{\iter \geq 1}$ all matched units and all the matched groups from all iterations $\iter$}

    {\bf Notation:} $\iter$: iterations, $D_{(\iter)}$ (resp. $D_{(\iter)}^m$)  = unmatched (resp. matched) units at the end of iteration $\iter$, $\mathcal{MG}_{(\iter)}$ = matched groups at the end of iteration $\iter$, $\Lambda_{(\iter)}$ = set of active covariate-sets at the end of iteration $\iter$ that are eligible to be dropped to form matched groups, $\Delta_{(\iter)}$ = set of covariate-sets  at the end of iteration $\iter$ that have been processed (i.e., have been considered to be dropped and for formulation of matched groups).
    
    {\bf Initialize: } $D_{(0)} = D, D_{(0)}^m = \emptyset, \mathcal{MG}_{(0)} = \emptyset, \Lambda_{(0)}=\{\{1\},...,\{p\}\}, \Delta_{(0)} = \emptyset,$
    $ \iter=1$ 

      \While{there is at least one treatment unit to match in $D_{(\iter-1)}$} {
       
          {\tt \small (find the `best' covariate-set to drop from the set of active covariate-sets)}
          
          Let $s_{(\iter)}^* \in \mathrm{arg}\max_{s \in \Lambda_{\iter-1}} \vv_s^T\mathbf{w}$ ($\vv_s \in \{0, 1\}^p$ denotes the indicator-vector of $s$ as in (\ref{equn:vector-s}))
          
          \If{early stopping condition is met}
          {\textbf{Exit while loop}}
                    
%
          
          $(D_{(\iter)}^{m}, \mathcal{MG}_{(\iter)}) = \BasicExactMatch(D, D_{(\iter-1)}, {\mathcal{J}}  \setminus s^*_{(\iter)})$ \texttt{\small (find matched units and main groups)}
          
     
         $Z_{(\iter)} =  \textrm{\GenerateNewActiveSets}(\Delta_{{(\iter-1)}}, s_{(\iter)}^*) $ (\texttt{\small generate new active covariate-sets})  

         $\Lambda_{(\iter)}=\Lambda_{{(\iter-1)}}\setminus \{s_{(\iter)}^*\}$ (\texttt{\small remove $s_{(\iter)}^*$ from the set of active sets}) 
                  
         $\Lambda_{(\iter)} = \Lambda_{{(\iter)}}\cup Z_{(\iter)} $ (\texttt{\small update the set of active sets})
         
         $\Delta_{(\iter)} = \Delta_{{(\iter-1)}}\cup \{s_{(\iter)}^*\}$ (\texttt{\small update the set of already processed covariate-sets})
         
          $D_{(\iter)} = D_{{(\iter-1)}}\setminus D_{{(\iter-1)}}^m$ \texttt{\small (remove matches)}
     
         $\iter = \iter+1$
         
     }
     \KwRet{$\{D_{(\iter)}^m, \mathcal{MG}_{(\iter)} \}_{\iter \geq 1}$}
     \label{algo:collapsingFLAME}
\end{algorithm}

Algorithm~\ref{algo:collapsingFLAME} gives the pseudocode of the \collapsingflame\ algorithm. It uses the monotonicity property stated in Proposition~\ref{prop:downward-closure}
and ideas from the \emph{apriori algorithm for association rule mining} \citep{Agrawal1994}. 
Instead of looping over all possible $2^{p}$ vectors to solve the AME, it considers a covariate-set $s$ for being dropped only if satisfies the monotonicity property of Proposition~\ref{prop:downward-closure}. For example, if  $\{1\}$ has been considered for being dropped to form matched groups, it would not process $\{1,2,3\}$ next because the monotonicity property requires  $\{1,2\}$, $\{1,3\}$, and $\{2,3\}$ to have been considered previously for being dropped. 	

The \collapsingflame\ algorithm uses the \BasicExactMatch\ (\emph{Grouped Matching with Replacement}) subroutine given in Algorithm \ref{algo:basicexactmatch} to form all valid main matched groups having at least one treated and one control unit. \BasicExactMatch\ takes a given subset of covariates and finds all subsets of treatment and control units that have identical values of those covariates. We use an efficient implementation of the {\tt group-by} operation in the algorithm from \cite{genericFLAME2017} that uses \emph{bit-vectors}. 
To keep track of main matched groups, 
\BasicExactMatch\ takes the entire set of units $D$ as well as the set of unmatched units from the previous iteration $D_{(\iter-1)}$ as input along with the covariate-set 
${\mathcal J} \setminus s_{(\iter)}^*$ to match on in this iteration. Instead of matching only the unmatched units in $D_{(\iter-1)}$ using the group-by procedure, it matches all units in $D$ to allow for matching with replacement as in the AME objective. It keeps track of the main matched groups for the unmatched units $D_{(\iter-1)}$.
\cut{
That is, for the units appearing in $D_{(\iter-1)}$, the returned matched groups are main matched groups since these units are matched for the first time. 
For all the other units, they may participate as auxiliary matched units to these same groups. If instead of \BasicExactMatch$^*$, the original \BasicExactMatch\ is invoked, the algorithm will perform matching without replacement. The pseudocode for  \BasicExactMatch$^*$ is very similar to \BasicExactMatch\ and is omitted. 
}

\begin{algorithm}
    \SetKwInOut{Input}{Input}
    \SetKwInOut{Output}{Output}
    \Input{Data $D$, unmatched Data $D^{um} \subseteq D =(X,Y,T)$, subset of indexes of covariates ${\mathcal J}^s \subseteq\{1,...,p\}$}
    \Output{Newly matched units $D^m$ using covariates indexed by ${\mathcal J}^s$ where groups have at least one treated and one control unit, 
    and main matched groups for $D^m$}
    $M_{raw}$ = 
    \texttt{group{-}by} $(D, {\mathcal J}^s)$ (\tt {\small form groups on $D$ by exact matching on $J^s$})\label{step:basic-1}\\ 
    $M$ = \texttt{prune}($M_{raw}$) ({\tt \small remove groups without at least one treatment and one control unit})\label{step:basic-2}\\
    $D^m$ = Subset of $D^{um}$ where the covariates match with some group in $M$ \texttt{\small (find newly matched units and their main matched groups)}\label{step:basic-3}\\
    
     \Return{$\{D^m,  M\}$ {\tt \small (newly matched units and main matched groups)}}
    \caption{Procedure {\BasicExactMatch}}
    \label{algo:basicexactmatch}
\end{algorithm}


\collapsingflame\ keeps track of two sets of covariate-sets: (1) The set of {\bf processed sets} $\Delta$ contains the covariate-sets whose main matched groups (if any exist) have already been formed. That is,  $\Delta$ contains $s$ if matches have been constructed on $\mathcal{J}\setminus s$ by calling the \BasicExactMatch\ procedure. (2) The set of {\bf active sets} $\Lambda$ contains the covariate-sets $s$ that are eligible to be dropped according to Proposition~\ref{prop:downward-closure}. For any iteration $\iter$, $\Lambda_{(\iter)} \cap \Delta_{(\iter)} = \emptyset$, i.e., the sets are disjoint, where $\Lambda_{(\iter)}, \Delta_{(\iter)}$ denote the states of $\Lambda, \Delta$ at the end of iteration $\iter$. Due to the monotonicity property stated in  Proposition~\ref{prop:downward-closure}, if $s \in \Lambda_{(\iter)}$, then each proper subset $r \subset s$ belonged to $\Lambda_{(\iter')}$ in an earlier iteration $\iter' < \iter$. Once an active set $s \in \Lambda_{(\iter-1)}$ is chosen as the optimal subset to drop (i.e., $s$ is $s_{(\iter)}^*$ in iteration $\iter$), $s$ is excluded from $\Lambda_{(\iter)}$ (it is no longer active) and is included in $\Delta_{(\iter)}$ as a processed set. In that sense, the active sets are generated and included in $\Lambda_{(\iter)}$ in a hierarchical manner similar to the apriori algorithm. A set $s$ is included in $\Lambda_{(\iter)}$ only if all of its proper subsets of one less size $r \subset s$, $|r| = |s| - 1$, have been processed. 

The procedure \GenerateNewActiveSets\ gives an efficient implementation of generation of new active sets in each iteration of \collapsingflame, and takes the currently processed sets $\Delta = \Delta_{(\iter-1)}$ and a newly processed set $s = s_{(\iter)}^*$ as input. Let $|s| = k$.
In this procedure, $\Delta^k \subseteq \Delta \cup \{s\}$ denotes the set of all processed covariate-sets in $\Delta$ of size $k$, and also includes $s$.
Inclusion of $s$ in $\Delta^k$ may lead to generation of 
a new active set $r$ of size $k+1$ only if all of $r$'s subsets of size $k$ (one less) have been previously processed. 
The new active sets triggered by inclusion of $s$ in $\Delta^k$ would be supersets $r$ of $s$ of size $k+1$ if all subsets $s' \subset r$ of size $|s'| = k$ belong to $\Delta^k$.  To generate such candidate supersets 
$r$, we can append $s$ with all covariates appearing in some covariate-set in $\Delta$ except those in $s$. However, this naive approach would iterate over many superfluous candidates for active sets. Instead, \GenerateNewActiveSets\  safely prunes some such candidates that cannot be valid active sets using {\bf support} of each covariate $e$ in $\Delta^k$, which is the number of sets in $\Delta^k$ containing $e$.  Indeed, for any covariate that is not frequent enough in $\Delta^k$, the monotonicity property ensures that any covariate-set that contains that covariate cannot be active. 
The following proposition shows that this pruning step does not eliminate any valid active set (proof is in the appendix):

\begin{proposition}\label{prop:newactiveset-correct}
If for a superset $r$ of a newly processed set $s$ where $|s| = k$ and $|r| = k+1$, all subsets $s'$ of $r$ of size $k$ have been processed (i.e. $r$ is  eligible to be active after $s$ is processed), then $r$ is included in the set $Z$ returned by \GenerateNewActiveSets.
\end{proposition}

The explicit verification step of whether all possible subsets of $r$ of one less size belongs to $\Delta^k$ is necessary, i.e., the above optimization only prunes some candidate sets that are guaranteed not to be active. For instance, consider $s = \{2, 3\}$,  $k = 2$, and $\Delta^2 = \{\{1,2\}, \{1,3\},\{3,5\}, \{5, 6\}\} \cup \{\{2, 3\}\}$. For the superset $r = \{2, 3, 5\}$ of $s$, all of $2, 3, 5$ have support of $\geq 2$ in $\Delta^2$, but this $r$ cannot become active yet, since the subset $\{2, 5\}$ of $r$ does not belong to $\Delta^2$.


Finally, the following theorem states the correctness of the \collapsingflame\ algorithm (proof is in the appendix).

\begin{theorem}\label{thm:main}
\textbf{(Correctness)}
The \collapsingflame\ algorithm 
solves the AME problem. 
\end{theorem}


Once the problem is solved, the main matched groups can be used to estimate treatment effects, by considering the difference in outcomes between treatment and control units in each group, and possibly smoothing the estimates from the matched groups to prevent overfitting of treatment effect estimates.

\begin{algorithm}[t!]
  \caption{Procedure {\it \GenerateNewActiveSets \label{alg:generatenewactivesets}} }
  \SetKwInOut{Input}{Input}
  \SetKwInOut{Output}{Output}  

    $1.$ \Input{  $s$ a newly dropped set of size $k$,\\ 
    $\Delta $ the set of previously processed sets }
    
    $2.$ {\bf Initialize: }   $Z$ = $\emptyset$ \texttt{\small (stores new active sets)}

    $3.$ $\Delta^k$ = $\{\delta \in \Delta \; \mid \; size(\delta) = k \} \cup \{s\}$     \texttt{\small (compute all subsets of $\Delta$ of size $k$ and also include $s$)}

    $4.$ $\Gamma$ = $\{\alpha \; \mid \; \alpha \in \delta$ and $\delta \in \Delta^k \}$     \texttt{\small (get all the covariates contained in sets in $\Delta^k $)} 
    
    $5.$ $\mathcal{S}_e$ = support of covariate $e$ in $\Delta^k$

    $6.$ $\Omega$ = $\{\alpha \; \mid \; \alpha \in \Gamma$ and $\mathcal{S}_\alpha \geq k \} \setminus s\ $ \texttt{ \small (get the covariates not in $s$ that have enough support)}

    $7.$ \If{$\{\forall e \in s : \mathcal{S}_e \geq k \}$      \texttt{\small (if all covariates in $s$ have enough support in $\Delta^k$)} }{
      

         $8.$ \For{ all $\alpha \in \Omega$  \texttt{  \small (generate new active set)} }{
          
            $9.$  $r = s \cup \{\alpha\} $

            $10.$  \If{all subsets $s' \subset r$, $|s'| = k$,  belong to $\Delta^k$ }{

             $11.$ add $r$ to $Z$ \texttt{\small (add newly active set $r$ to $Z$)}
              } 
           }   

      }
   $12.$ \KwRet{$Z$}
     
     \vspace{3mm} 
   
   \vspace{3mm}
   
{\bf Example (follow line number correspondence)} \\

     $1.$ $s = \{2,3\}$, $k=2$, \\ 
    $ \Delta = \{ \{1\}, \{2\}, \{3\}, \{5\}, \{1,2\},
     \{1,3\}, \{1, 5\} \}$ 
    
   $2.$  $Z$ = $\emptyset$ 
    
   $3.$ $\Delta^2$ = $\{\{1,2\},\{1,3\},\{2,3\}, \{1, 5\} \}$

   $4.$  $\Gamma$ = $\{1,2,3, 5 \}$ 
   
     $5.$ $\mathcal{S}_1 = 3, \mathcal{S}_2 = 2, \mathcal{S}_3 = 2, \mathcal{S}_5 = 1$

   $6.$  $\Omega$ = $\{1,2,3 \} \setminus \{2,3\} = \{1\}$

     $7.$  $True:$ both 1 and 2 have support $\geq 2$
     
       $8.$ $\alpha = 1 $ (only one value)
                   
         $9.$  $r = \{2,3\} \cup \{1\} = \{1,2,3\}$
          
         $10.$  $True$ (subsets of $r$ of size 2 are $\{1,2\},\{1,3\},\{2,3\})$

          $11.$    $ Z = \{\{1,2,3\}\}$ 

 
     $12.$  \KwRet{ $Z = \{\{1,2,3\}\}$}


\end{algorithm}

\section{Almost Matching Exactly with Adaptive Weights}\label{sec:adaptive}
We now generalize the AME framework so the weights are adjusted adaptively for each subset of variables. The weights are chosen using machine learning on a hold-out training set. Let us consider a trivial variation of the AME problem with fixed weights and then generalize it to handle adaptive weights. 

\noindent \textbf{Almost Matching Exactly with Fixed Weights, Revisited:} We will use squared rewards $w_j^2$ this time. 
For a given treatment unit $u$ with covariates $\x_u$, compute the following, which is the maximum sum of rewards $\{w_j^2\}_{j=1,..,p}$ we can attain for a valid matched group (that contains at least one  control unit):
\begin{eqnarray}\nonumber
\lefteqn{\btheta^{u*}\in \textrm{argmax}_{\btheta\in\{0,1\}^p}
\btheta^T(\w\circ\w) ~~\textrm{ s.t. } }\\
&\exists \ell \textrm{ with } T_\ell=0 \textrm{ and } \x_{\ell}\circ \btheta = \x_u\circ \btheta.&\label{eqn:special}
\end{eqnarray}
The solution to this is an indicator of the optimal set of covariates to match unit $u$ on.
For treatment unit $u$, again, the \textbf{main matched group} for $u$ contains all units $\ell$ so that $\x_u\circ \vv^{u*} = \x_{\ell}\circ \vv^{u*}$. Now we provide the (more general) adaptive version of AME. 


\textbf{Full Almost Matching Exactly (Full-AME):}  Denote $\btheta\in\{0,1\}^p$ as an indicator vector for a subset of covariates to match on. Define the matched group for unit $u$ with respect to covariates $\btheta$ as the units that match $u$ exactly on the covariates $\btheta$:
$$\vspace{-.15cm}\MG_{\btheta}(u) = \{v: \x_v \circ \btheta = \x_u \circ\btheta \}.
$$
The usefulness of a set of covariates $\btheta$ is now determined by \textit{how well they can be used together to make out-of-sample predictions}. 
Specifically, the prediction error $\PE(\theta)$ is defined with respect to a class of functions $\mathcal{F}$ as:
$
\PE_{\mathcal{F}} (\btheta) = \min_{f\in\mathcal{F}} \mathbb{E} (f(X \circ \btheta ,T)-Y)^2,
$
where the expectation is taken over $X$, $T$ and $Y$. Its empirical counterpart is defined with respect to a separate random sample from the distribution, used as a training set $\{\x_i^{tr},T_i^{tr},y_i^{tr}\}_{i \in \textrm{ training}}$, specifically: 
\begin{equation*}
\widehat{\PE}_{\mathcal{F}} (\btheta) = \min_{f\in\mathcal{F}} \sum_{i \in \textrm{ training}} (f(\x_i^{tr} \circ \btheta ,T^{tr}_i)-y_i^{tr})^2. \label{eq:pehat}
\end{equation*}
The training set is only used to calculate prediction error, not for matching.
Using this, the best prediction error we could hope to achieve for a nontrivial matched group containing treatment unit $u$ uses the following covariates for matching:
\begin{eqnarray*}
\btheta_u^{*} \in \mathrm{arg}\min_{\theta}\widehat{\PE}_\mathcal{F} (\btheta)  &\textrm{s.t.}&\exists \ell\in \MG_{\btheta}(u) \textrm{ where } T_{\ell}=0
\end{eqnarray*}
The \textbf{main matched group} for $u$ is defined as $\MG_{\btheta^*_u}(u)$.
The goal of the Full-AME problem is to find the main matched group $\MG_{\btheta^*_u}(u)$ for all units $u$. 

The class of functions $\mathcal{F}$ can include nonlinear functions. We can use variable importance measures for prediction on $\widehat\PE_{\mathcal{F}}$ such as permutation importance (also called model reliance) to determine the variable's weight. If $\mathcal{F}$ includes linear models, the weight $w_j$ for feature $j$ would be the absolute value of feature $j$'s coefficient.

The Full-AME problem reduces to the fixed-squared-weights version under specific conditions, such as when $\mathcal{F}$ is a single function $f$, which is linear with fixed linear weights ($\w, w_T$) and $f(\x\circ\btheta, T)=(\w\circ\btheta)^T  (\x\circ\btheta) + w_T T$,
where $\w$ is the ground-truth coefficient vector that generates $y$, and $\widehat{\PE}_{\btheta}$ is determined by the sum of $w_j^2$ weights for covariates determined by the feature-selector vector $\btheta$. This reduction is discussed formally by \cite{genericFLAME2017}.

In order to solve Full-AME, a step is needed in Algorithm \ref{algo:collapsingFLAME} at the top of the while loop that updates the weights for each covariate-set we could choose at that iteration. 
In particular, we let $$s_{(\iter)}^*\in\arg\min_{s\in{\Lambda_{(h-1)}}}\widehat\PE(\theta_s),$$ where $\Lambda_{(h-1)}$ is the active set of covariates, and the predictive error is computed over the training set with respect to a pre-specified class of models, $\mathcal{F}$. In the implementation in this paper we consider linear functions fit separately on the treated and the control units in the training set using ridge regression (that is, we add a ridge penalty to Eq~\eqref{eq:pehat}). 


\subsection{Early Stopping of \collapsingflame}
It is important that \collapsingflame\ be \textit{stopped early} when the quality of matches produced falls. 
In dropping covariates, its prediction error $\widehat{\PE}_{\mathcal{F}}$ should never increase too far above its original value using all the covariates. This ensures the quality of every matched group: the covariates $\btheta^*_u$ for every matched group thus obey $\widehat{\PE}_\mathcal{F}(\btheta^*_u)<\min_{\btheta}\widehat{\PE}_\mathcal{F}(\btheta) +\epsilon$, where the choice of $\epsilon$ (perhaps 5\%) determines stopping. As such the while loop in Algorithm~\ref{algo:collapsingFLAME} should not only check whether there are more units to match, but also whether the predictive error has increased too much.


\subsection{Hybrid FLAME-DAME}
\label{sec:hybrid}
The \collapsingflame\ algorithm solves the Full-AME problem, whereas FLAME \citep{genericFLAME2017} approximates its solution. This is because FLAME uses backwards feature selection, whereas \collapsingflame\ calculates the solution without approximation. For problems with many features, we can use FLAME to remove the less relevant features, and then switch to \collapsingflame\ when we start to remove some of the more influential features. This hybrid algorithm scales substantially better, possibly without any noticeable loss in the quality of matches. 

Matching-after-learning-to-stretch (MALTS) \citep{ParikhRuVo18} has been combined with FLAME and \collapsingflame\ to handle mixed real and categorical covariates.

\subsection{Other Estimands}
While CATEs are the most granular estimands, aggregate estimands such as Average Treatment Effect (ATE) and Average Treatment Effect on the Treated (ATT) may be of interest. Since \collapsingflame\ matches with replacement,
standard techniques (e.g., frequency weights) should be used \citep{stuart2010matching,abadie2004implementing}.

\section{SIMULATIONS}
\label{sec:Sim}

We present results under several data generating processes.
We show that \collapsingflame\ produces higher quality matches than popular matching methods such as 1-PSNNM (propensity score nearest neighbor matching) and Mahalanobis distance nearest neighbor matching, and better treatment effect estimates than black box machine learning methods such as Causal Forest (which is not a matching method, and is not interpretable). 
The `MatchIt' R-package \citep{matchit} was used to perform 1-PSNNM and Mahalanobis distance nearest neighbor matching (`Mahalanobis'). For Causal Forest, we used the `grf' R-package \citep{grf2016}. \collapsingflame\ also improves over FLAME \citep{genericFLAME2017} with regards to the quality of matches. Other matching methods (optmatch, cardinality match) do not scale to large problems and thus needed to be omitted. 

Throughout this section, the outcome is generated with $y = \sum_{i}^{} \alpha_i x_i + T \sum_{i=1}^{} \beta_i x_i + T\cdot U \sum_{i,\gamma,\gamma>i}  x_i x_{\gamma}$
where $T \in \{ 0, 1\}$ is the binary treatment indicator. This generation process includes a baseline linear effect, linear treatment effect, and quadratic (nonlinear) treatment effect. We vary the distribution of covariates, coefficients ($\alpha$'s, $\beta$'s, $U$), and the fraction of treated units. We report conditional average treatment effects on the treated.

\subsection{Presence of Irrelevant Covariates}\label{sec:irrel}
\begin{figure*}[]
\centering
\includegraphics[width=\linewidth]{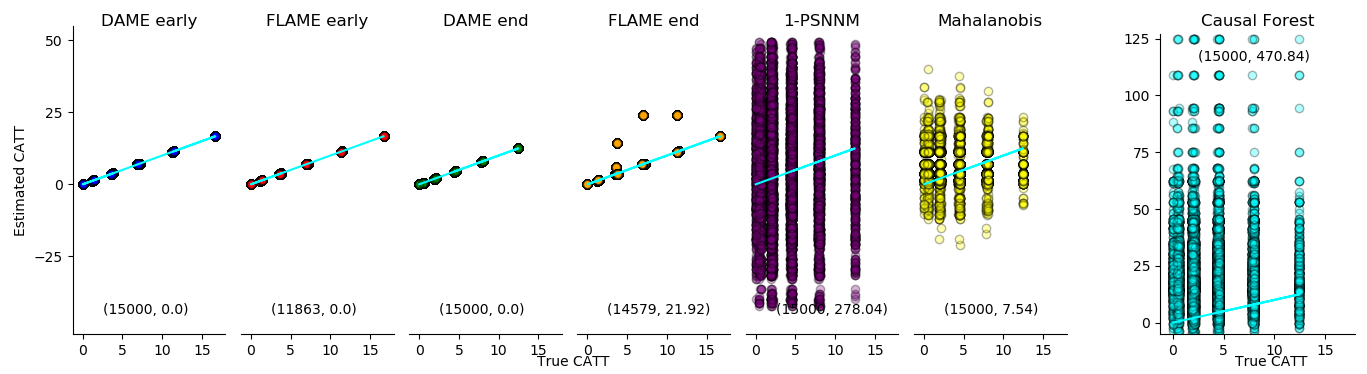}
\caption{\small Estimated CATT vs$.$ True CATT (Conditional Average Treatment Effect on the Treated). \collapsingflame\ and FLAME perfectly estimate the CATTs before dropping important covariates. \collapsingflame\ matches all units without dropping important covariates, but FLAME needs to stop early in order to avoid poor matches. 
All other methods are sensitive to irrelevant covariates and give poor estimates. The two numbers on each plot are the number of matched units and MSE.\label{fig:irrelevant} \normalsize}
\end{figure*}

A basic sanity check for matching algorithms is how sensitive they are to irrelevant covariates. To that end, we run experiments with a majority of the covariates being irrelevant to the outcome. 
For important covariates $1\le i \le 5$ let $\alpha_i \sim N(10s, 1) $  with $s \sim \text{Uniform}\{-1, 1 \}$, $\beta_i \sim N(1.5, 0.15)$, 
$x_i \sim \text{Bernoulli}(0.5)$. 
For unimportant covariates $5 < i \le 15$, $x_i \sim \text{Bernoulli}(0.1)$ in the control group and $x_i \sim \text{Bernoulli}(0.9)$ in the treatment group so there is little overlap between treatment and control distributions. 
This simulation generates 15000 control units, 15000 treatment units, 5 important covariates and 10 irrelevant covariates. 
{\bf Results:} In Figure~\ref{fig:irrelevant},  \collapsingflame\ (even with early stopping) runs to the end and matches on all units because the stopping criteria is never met. In this figure, \collapsingflame\ finds all high-quality matches even after important covariates are dropped.  In contrast, FLAME achieves the optimal result before dropping any important covariates and generates some poor matches after dropping important covariates.
 However, even FLAME's worst case scenario is better than the comparative methods, all of which perform poorly in the presence of irrelevant covariates. Causal Forest is especially ill suited for this case.

\subsection{Exponentially Decaying Covariates}
\begin{figure}[]
\centering
\includegraphics[width=.48\textwidth,]{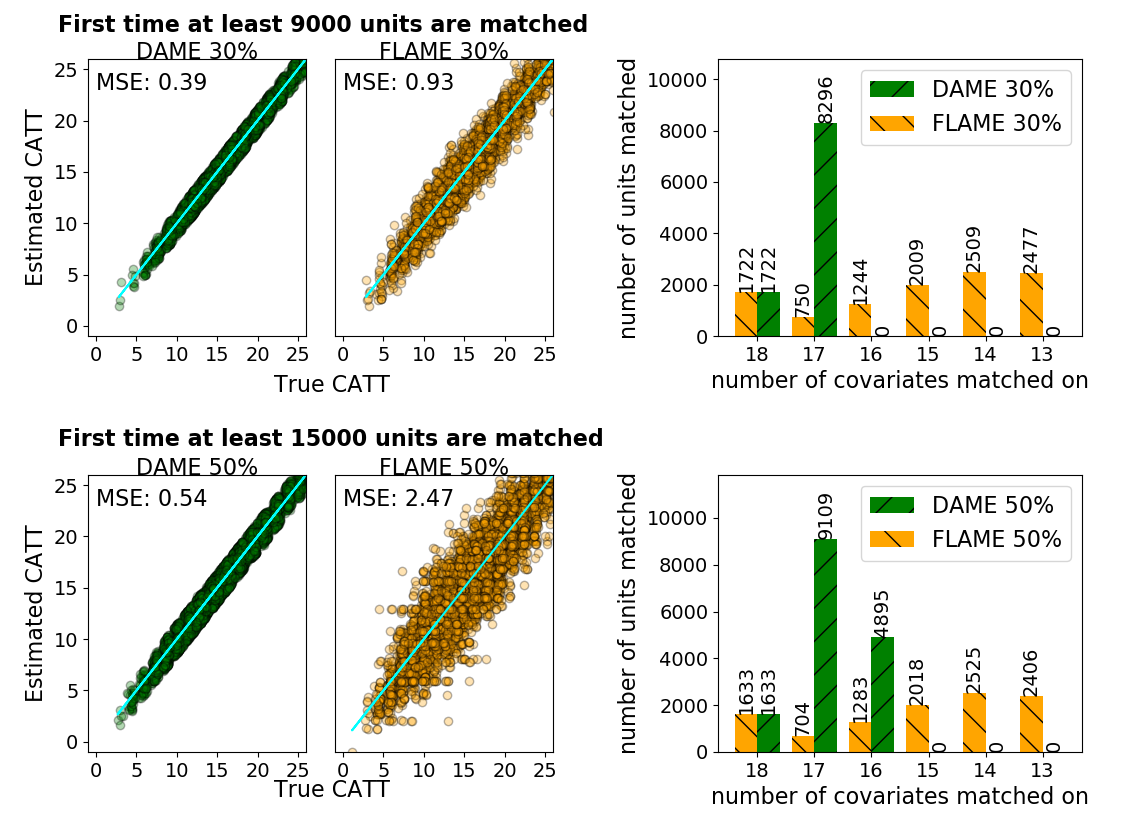}
\caption{\small
\collapsingflame\ makes higher quality matches early on. Rows correspond to stopping thresholds (top row 30\%, bottom row 50\%).
\collapsingflame\ matches on more covariates than FLAME, yielding lower MSE from matched groups. \label{fig:GENvsCOL}}
\end{figure}

An advantage of \collapsingflame\ over FLAME is that it produces more high quality matches before resorting to lower quality matches.  To test this, we considered covariates of decaying importance, letting the $\alpha$ parameters decrease exponentially as $\alpha_i = 64 \times \left( \frac{1}{2} \right)^i$. 
We evaluated performance when $\approx 30\%$ and $50\%$ of units were matched. {\bf Results:} As Figure~\ref{fig:GENvsCOL} shows,  \collapsingflame\  matches on more covariates, yielding better estimates than FLAME. 


\subsection{Imbalanced Data}\label{sec:imb}
Imbalance is common in observational studies: there are often substantially more control than treatment units. 
The data for this experiment has covariates with decreasing importance. A fixed batch of 2000 treatment and 40000 control units were generated. We sampled from the controls to construct different imbalance ratios: 40000 in the most imbalanced case (Ratio 1), then 20000 (Ratio 2), and 10000 (Ratio 3). 
{\bf Results:}  
Table \ref{tab:mse_imb} reveals that FLAME and \collapsingflame\ outperform the nearest neighbor matching methods. 
\collapsingflame\ is distinctly better than FLAME. Additionally, \collapsingflame\ has an average of 4 covariates not matched on, with $\approx 84\%$ of units matched on all but 2 covariates, 
whereas FLAME averages 7 covariates not matched on and only $\approx 25\%$ units matched on all but 2 covariates. Detailed results are in the longer version \citep{LongerVersion}.

\begin{table}[]
\setlength{\tabcolsep}{5pt}
\centering
\caption{MSE for different imbalance ratios}
\small{
\begin{tabular}{@{} l c|c|c @{}} 
\toprule
   &  \multicolumn{3}{c @{}}{ Mean Squared Error (MSE) } \\ 
   & Ratio 1 & Ratio 2  & Ratio 3 \\ 
\cmidrule(l){1-4}
\midrule
\collapsingflame     & {\bf 0.47} & {\bf 0.83} & {\bf 1.39}  \\
FLAME     & 0.52 & 0.88 & 1.55  \\
Mahalanobis    & 26.04 & 48.65 & 64.80  \\
1-PSNNM      & 246.08 & 304.06 & 278.87 \\
\bottomrule
\end{tabular}
}
\label{tab:mse_imb}
\end{table}


\begin{figure}[]
\centering
\includegraphics[width=0.48\textwidth 
]{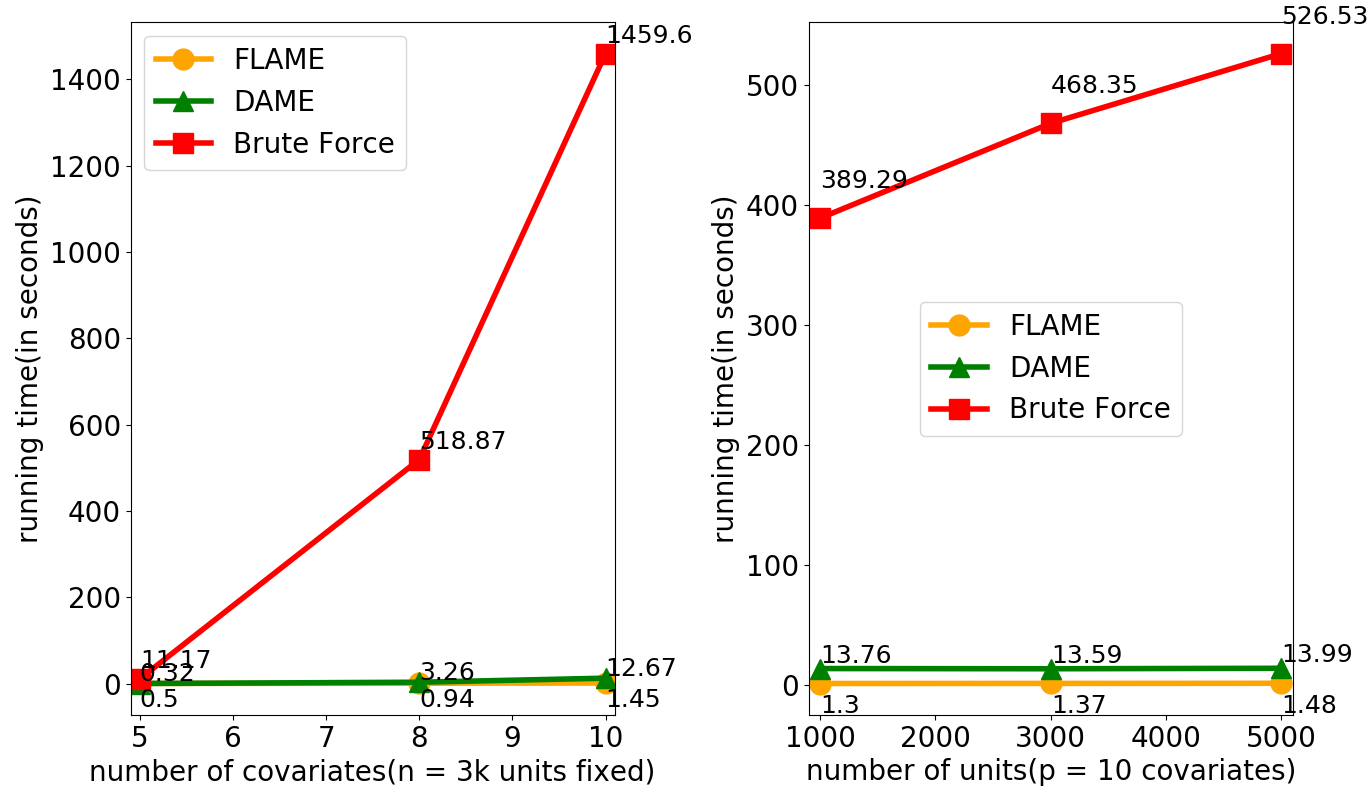}
\caption{\small Run-time comparison between \collapsingflame\, FLAME, and brute force. \textit{Left:} varying number of units. \textit{Right:} varying number of covariates. \label{fig:runtime}}
\end{figure}

\subsection{Run Time Evaluation}\label{sec:runtime}
We compare the run time of \collapsingflame\ with a brute force solution (AME Solution 1 described in Section~\ref{sec:AEM}). All experiments were run on an Ubuntu 16.04.01 system with Intel Core i7 Processor (Cores: 8, Speed: 3.6 GHz), 8 GB RAM.
{\bf Results:} As shown in Figure~\ref{fig:runtime}, FLAME provides the best run-time performance because it incrementally reduces the number of covariates, rather than solving Full-AME. 
On the other hand, as shown in the previous simulations, \collapsingflame\ produces high quality matches that the other methods do not. It solves the AME much faster than brute force. The run time for \collapsingflame\ could be further optimized through simple parallelization of the checking of active sets.

\subsection{Missing Data} 
Missing data problems are complicated in matching. Normally one would impute missing values, but matches become less interpretable when matching on imputed values. If we match only on the raw values, \collapsingflame\ has an advantage over FLAME because it can simply match on as many non-missing relevant covariates as possible. When data are imputed, \collapsingflame\ still maintains an advantage over FLAME, possibly because it can match on more raw covariate values and fewer imputed values. Details are in the Appendix~\ref{exp:miss_data}.

\subsection{Effect of Noise}
In Appendix~\ref{exp:noise}, we study how \collapsingflame\ performs in the presence of noise. 
In particular, \collapsingflame\ tends to outperform FLAME in the presence of noise. 

\section{BREAKING THE CYCLE OF DRUGS AND CRIME}
Breaking The Cycle (BTC) \citep{btc2006} is a social program conducted in several U.S. states designed to reduce criminal involvement and substance abuse among current offenders. 
We study the effect of participating in the program on reducing non-drug future arrest rates.
The details of the data and our results are in Appendix \ref{sec:btc-details}. We compared CATE predictions of \collapsingflame\ and FLAME to double check the performance of a black box support vector machine (SVM) approach that predicts positive, neutral, or negative treatment effect for each individual. The result is that \collapsingflame\ and the SVM approach agreed on most of the exactly matched units. All of the units for which exact matching predicted approximately zero treatment effect all have a ``neutral'' treatment effect predicted label from the SVM. Most other predictions were similar between the two methods. There were only few disagreements between the methods. Upon further investigation, we found that the differences are due to the fact that \collapsingflame\ is a matching method and not a modeling method; the estimates could be smoothed afterwards if desired to create a model. In particular, one of the two disagreeing predictions between the SVM and \collapsingflame\ has a positive treatment CATE prediction, but it was closer in Hamming distance to units predicted to have negative treatment effects. With smoothing, its predicted CATE may have also become negative.

 
 

\section{CONCLUSIONS}\label{sec:conclusions}

\collapsingflame\ produces matches that are of high quality. Its estimates of individualized treatment effects are as good as the (black box) machine learning methods we have tried.
Other methods can match individuals together whose covariates look nothing alike, whereas the matches from \collapsingflame\ are interpretable and meaningful, because they are almost exact; units are matched on covariates that together can be used to predict outcomes accurately.
Code is publicly available at: \url{https://github.com/almost-matching-exactly/DAME} .

\subsubsection*{Acknowledgements:} 
This work was supported in part by NIH award 1R01EB025021-01, NSF awards IIS-1552538 and IIS-1703431, a
DARPA award under the L2M program, and a Duke University Energy Initiative Energy Research Seed Fund (ERSF).

\appendix
\title{Interpretable Almost Matching Exactly for Causal Inference\\
{\bf Supplementary Material}}



\author{
 Yameng Liu,  Awa Dieng, 
  Sudeepa Roy, Cynthia Rudin, Alexander Volfovsky
}

\date{}
\pagenumbering{gobble}
\maketitle
\begin{center}
{\Huge \textbf{Appendix}}
\end{center}
\section{Na\"ive AME solutions}\label{sec:naive-AMEr}
In this section we present the complete outline of the two straightforward (but inefficient) solution to the AME problem (described in Section~\ref{sec:AEM}) for all units. 

\textbf{AME Solution 1 (quadratic in $n$, linear in $p$):} For all treatment units $t$, we (i) iterate over all control units $c$, (ii) find the vector $\vv_{tc} \in \{0, 1\}^p$ where $\vv_{tcj}=1$ if $t$ and $c$ match on covariate $j$ and 0 otherwise, (iii) find the control unit(s) with the highest value of $\vv_{tc}^T\mathbf{w}$, and (iv) return them as the main matched group for the treatment unit $t$. 
Repeat the same procedure for each control unit $c$. Note that the CATE for each unit is computed based on its main matched group which means that the outcome of each unit can contribute to the computation of CATEs for multiple units.
\cut{Whenever a previously matched unit $\alpha$ is matched to a previously unmatched unit $\eta$, record $\eta$'s main matched group as an auxiliary group for the previously matched unit $\alpha$. When all units are `done' (all units are either matched already or cannot be matched) then stop, and compute the CATE for each treatment and control unit using its main matched group. If a unit belongs to auxiliary matched groups then its outcome is used for computing both its own CATE (in its own main matched group) and the CATEs of units for whom it is in an auxiliary group (e.g., $\alpha$ will be used to compute $\eta$'s estimated CATE).}
This algorithm is polynomial in both $n$ and $p$, however, the quadratic time complexity in $n$ also makes this approach impractical for large datasets (for instance, when we have more than a million units with half being treatment units).\\

\textbf{AME Solution 2 (order $n \log n$, exponential in $p$:)} This approach solves the AME problem simultaneously for all treatment and control units for a fixed weight vector $\mathbf{w}$. First, (i) enumerate every $\vv\in\{0,1\}^p$ (which serves as an indicator for a subset of covariates), (ii) order the $\vv$'s according to  $\vv^T \mathbf{w}$, (iii) call \BasicExactMatch\ for every $\vv$ in the predetermined order, (iv) the first time each unit is matched during a \BasicExactMatch\ procedure, mark that unit with a `done' flag, and record its corresponding main matched group and
compute the CATE for each treatment and control unit using its main matched group. Each unit's outcome will be used to estimate CATEs for every auxiliary group that it is a member of, as before.
Although this approach can use an efficient `group by' function (e.g., an implementation using \emph{bit-vectors} or \emph{database/SQL queries} as discussed by \cite{genericFLAME2017}), which can be implemented in $O(n \log n)$ time by sorting the units, iterating over all possible vectors $\vv\in\{0,1\}^p$ makes this approach unsuitable for practical purposes (exponential in $p$).


\section{Proof of Proposition~\ref{prop:newactiveset-correct}}

\textbf{Proposition~\ref{prop:newactiveset-correct}} \emph{
If for a  superset $r$ of a newly processed set $s$ where $|s| = k$ and $|r| = k+1$, all subsets $s'$ of $r$ of size $k$ have been processed (i.e. $r$ is  eligible to be active after $s$ is processed), then $r$ is included in the set $Z$ returned by \GenerateNewActiveSets.
}

\begin{proof}
Suppose all subsets of $r$ of size $k$ are already processed and belong to $\Delta^k$. Let $f$ be the covariate in $r \setminus s$. Clearly, $f$ would appear in $\Delta^k$, since at least one subset $s' \neq s$ of $r$ of size $k$ would contain $f$, and $s' \in \Delta^k$. Further all covariates in $r$, including $f$ and those in $s$ will have support at least $k$ in $\Delta^k$. To see this, note that there are $k+1$ subsets of $r$ of size $k$, and each covariate in $r$ appears in exactly $k$ of them. Hence $f \in \Omega$, which the set of high support covariates. Further, the `if' condition to check minimum support for all covariates in $s$ is also satisfied. In addition, the final `if' condition to eliminate false positives is satisfied too by assumption (that all subsets of $r$ are already processed). Therefore $r$ will be included in $Z$ returned by the procedure.
\end{proof}

\section{Proof of Theorem~\ref{thm:main}}

\textbf{Theorem~\ref{thm:main}} \emph{\textbf{(Correctness)}
The \collapsingflame\ algorithm 
solves the AME problem. }

\begin{proof}
Consider any treatment unit $t$. Let $s$ be the set of covariates in its main matched group returned in \collapsingflame\ (the while loop in \collapsingflame\ runs as long as there is a treated unit and the stopping criteria have not been met, and the \BasicExactMatch\ returns the main matched group for every unit when it is matched for the first time). Let $\vv_s$ be the indicator vector of $s$ (see Eq.~\ref{equn:vector-s}). Since the \BasicExactMatch\ procedure returns a main matched group only if it is a \emph{valid} matched group containing at least one treated and one control unit (see Algorithm~\ref{algo:basicexactmatch}), and since all units in the matched group on $s$ have the same value of covariates in 
$\mathcal{J} \setminus s$, there exists a unit $\ell$ with $T_\ell = 0$ and $\x_{\ell} \circ \vv_s = \x_t\circ\vv_s$. 

Hence it remains to show that the covariate set $s$ in the main matched group for $t$ corresponds to the maximum weight $\vv^T\mathbf{w}$ over all $\vv$ for which there is a valid matched group. Assume that there exists another covariate-set $r$ such that $\vv_r^T \mathbf{w} > \vv_s^T \mathbf{w}$, there exists a unit $\ell'$ with $T_{\ell'} = 0$ and $\x_{\ell'} \circ \vv_r = \x_t\circ\vv_r$, and gives the maximum weight $\vv_r^T \mathbf{w}$ over all such $r$. Then,
\begin{itemize}
    \itemsep0em
    \item[(i)]$r$ cannot be a (strict) subset of $s$, since \collapsingflame\ ensures that all subsets are processed before a superset is processed to satisfy the downward closure property in Proposition~\ref{prop:downward-closure}.
    \item[(ii)] $r$ cannot be a (strict) superset of $s$. Recall that $\theta_{s,j}$ is 1 for covariates $j$ that are not in $s$ (analogously for $r$). If $r$ is a strict superset of $s$, then we would have $\vv_r^T \mathbf{w} \leq \vv_s^T \mathbf{w}$, which violates the assumption that $\vv_r^T \mathbf{w} > \vv_s^T \mathbf{w}$ for non-negative weights.
\end{itemize}
Given (i) and (ii), $r$ and $s$ must be incomparable (there exist covariates in both $r \setminus s$ and $s \setminus r$). 
Suppose the active set $s$ was chosen in iteration $\iter$. If $r$ was processed in an earlier iteration $\iter' < \iter$, since $r$ forms a valid matched group for $t$, 
it would give the main matched group for $t$, violating the assumption that $s$ was chosen by \collapsingflame\ to form the main matched group for $t$, rather than $r$.

Next, we argue that 
$r$ must be active at the start of iteration $\iter$, and will be chosen as the best covariate set in iteration $\iter$, leading to a contradiction.



Note that we start with all singleton sets as active sets in $\Lambda_{(0)} = \{\{1\}, \cdots, \{p\}\}$ in the \collapsingflame\ algorithm. Consider any singleton subset $r_0 \subseteq r$ (comprising a single covariate in $r$). Due to the downward closure property in Proposition~\ref{prop:downward-closure},   $\vv_{r_0}^T \mathbf{w} \geq \vv_r^T \mathbf{w}
> \vv_s^T \mathbf{w}$. Hence all of the singleton subsets of $r$ will be processed in earlier iterations $\iter' < \iter$, and will belong to the set of processed covariate sets $\Delta_{(\iter-1)}$. 

Repeating the above argument, consider any subset $r' \subseteq r$. It holds that $\vv_{r'}^T \mathbf{w} \geq \vv_r^T \mathbf{w} > \vv_s^T \mathbf{w}$. All subsets $r'$ of $r$ will be processed in earlier iterations $\iter' < \iter$ starting with the singleton subsets of $r$. In particular, all subsets of size $|r|-1$ will belong to $\Delta_{(\iter-1)}$. As soon as the last of those subsets is processed, the procedure \GenerateNewActiveSets\ will include $r$ in the set of active sets in a previous iteration $\iter' < \iter$. Hence if $r$ is not processed in an earlier iteration, it must be active at the start of iteration $\iter$, leading to a contradiction. 

Hence for all treatment units $t$, the covariate-set $r$ giving the maximum value of $\vv_r^T \mathbf{w}$ will be used to form the main matched group of $t$, showing the correctness of the \collapsingflame\ algorithm.  
\end{proof}


\section{Additional simulations and results}\label{exp:app}

\subsection{Missing Data}\label{exp:miss_data}



In this section, we consider the case when data are missing at random. We compared performance of FLAME and \collapsingflame\ both with and without multiple imputation on missing data. 

To allow for missing values in \collapsingflame\ and FLAME we randomly construct a $n$(number of units) $\times$ $p$(number of covariates) binary matrix $O$  where $o_{ij}=1$ if covariate $j$ is unobserved (``missing'') for unit $i$. Temporarily setting ``missing'' as just another category for each variable, we proceed with the algorithm by adding a condition for a matched group to be valid: if covariates $J' = \{j_1,\dots,j_{p'}\}$ are being matched on, then $\sum_{j\in J'} o_{ij}=0$ should hold for each unit $i$ in the group. A greater than $0$ sum means that unit $i$ has missing values in $J'$ to be matched on, and this matched group thus is invalid -- unit $i$ must then be removed from the matched group; we handled missing values in this way. For imputation, we used the Multiple Imputation Chained Equations algorithm in the `mice' R package, constructed 10 multiply-imputed datasets and matched on the imputed values. Estimates from each of the multiply imputed datasets were then combined using Rubin's rules. 

\begin{figure}[h!]
\centering
\includegraphics[width=.45\textwidth]{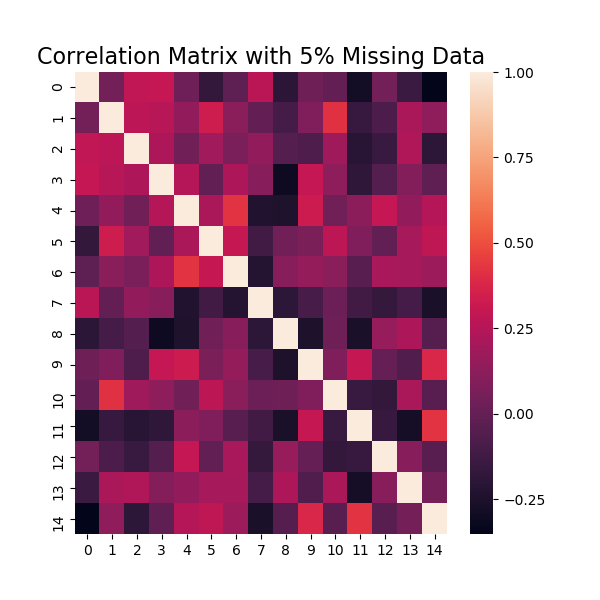}
\vspace{.3in}
\caption{\small Correlation matrix that defines the relationship among the covariates $X$.\label{fig:corr}}
\end{figure}

We generate covariates $\x$ by first sampling $\mathbf{z}\sim N_p(\mu,\Sigma)$, where $\mu$ is $0$ and covariates are correlated, meaning that $\Sigma$ is not the identity matrix ($\Sigma$ is plotted in Figure~\ref{fig:corr}.) We then let $x_j=1_{z_j>0}$. This leads to a correlation among the covariates $X$.  
We generated 1000 control and 1000 treated units; 5\% of the data are missing at random.
\begin{figure}[ht]
\centering
\includegraphics[width=.5\textwidth]{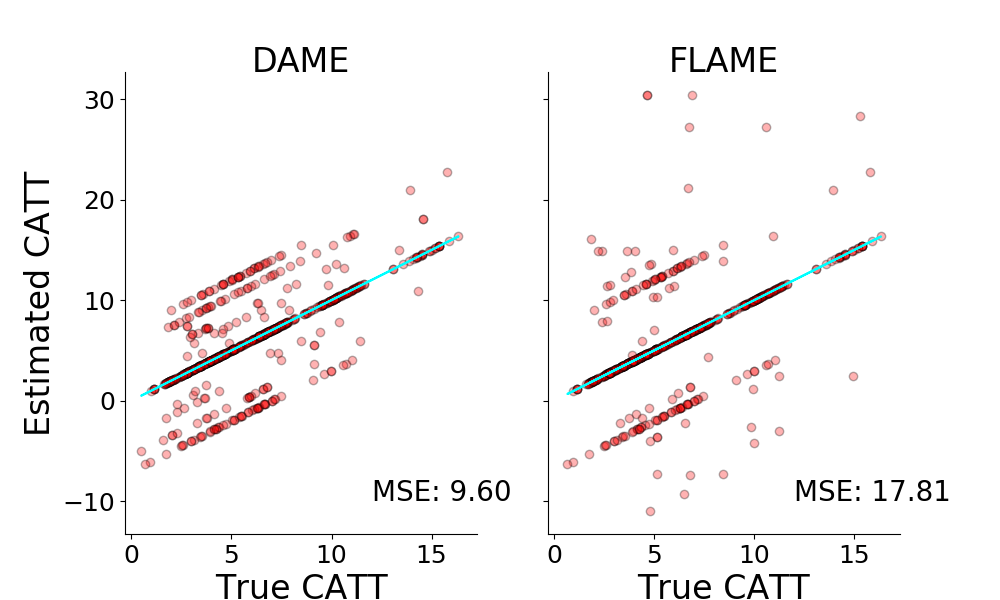}
\caption{\small Without Imputation: Comparison of true CATE and CATE estimates for \collapsingflame\ (left) and FLAME (right) without imputation on the missing data experiment.
  \label{fig:noimpute}}
\end{figure}

\begin{figure}[ht]
\centering
\includegraphics[width=.45\textwidth]{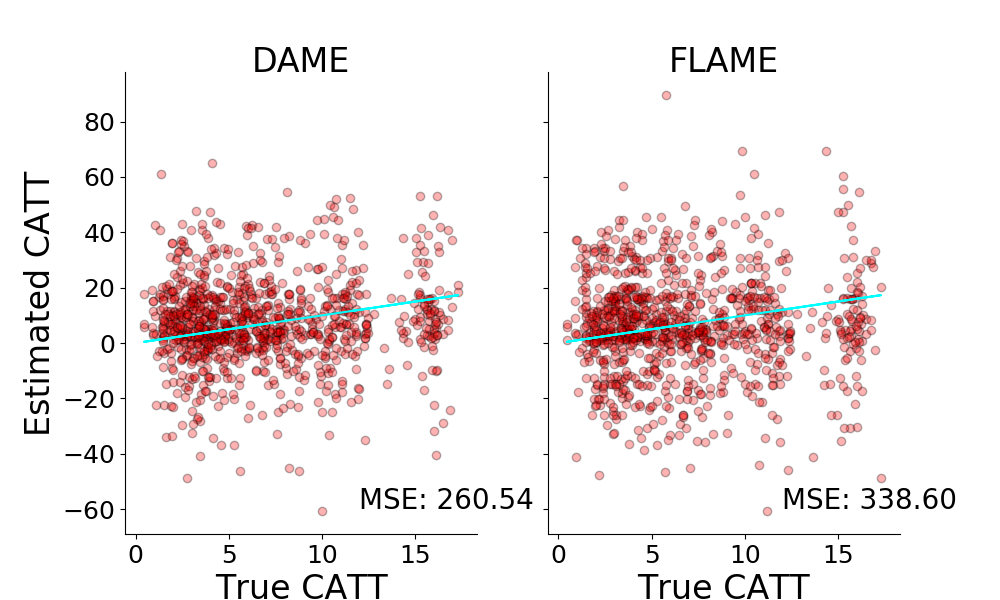}
\caption{Comparison of true CATE and CATE estimates with imputation for \collapsingflame\ (left) and FLAME (right).
\label{fig:impute2}}
\end{figure}

Figure~\ref{fig:noimpute} contains the results without imputation, and Figure~\ref{fig:impute2} contains the results with imputation. Either with imputation or without imputation, \collapsingflame\ outperforms FLAME in terms of CATT estimation. The MSE of \collapsingflame\ without imputation is 9.60 whereas FLAME's MSE without imputation is 17.81. Figure~\ref{fig:impute2} shows the comparison with imputation. With imputation, the MSE of \collapsingflame\ is 260.54 whereas FLAME's MSE is 338.60. The MSEs are larger in the experiments with imputation because by dropping covariates, both \collapsingflame\ and FLAME match on more units than without imputation, which generates larger bias in the estimation. 

\subsection{Effect of Noise}\label{exp:noise}
In this simulation we study the effects of measurement noise on the performance of \collapsingflame. We use 15,000 treated units, 15,000 control units, 10 important covariates and 5 unimportant covariates.
We change our generative process in the following way to add noise:
\begin{equation}\label{equn:model with noise}
y = \alpha_0 + \sum_{i}\alpha_{i}x_{i} + T\sum_{i=1}\beta_{i}x_{i} + T\cdot U\sum_{i,\gamma,\gamma > i}x_{i} x_{\gamma} + \tau\epsilon,
\end{equation}
where $U$ is 1, and for important covariates $1\le i \le 10$ let $\alpha_i \sim N(10s, 1) $  with $s \sim \text{Uniform}\{-1, 1 \}$, $\beta_i \sim N(1.5, 0.15)$, $x_i \sim \text{Bernoulli}(0.5)$. 
For unimportant covariates $10 < i \le 15$, $x_i \sim \text{Bernoulli}(0.1)$ in the control group and $x_i \sim \text{Bernoulli}(0.9)$ in the treatment group so there is little overlap between treatment and control distributions. $\epsilon$ denotes noise chosen as either $\epsilon_1$, $\epsilon_2$ and $\epsilon_3$ below:\\
Noise 1: $\epsilon_1 \sim N(mean=1, sd=0.2) $ \\
Noise 2: $\epsilon_2 \sim N(mean=1, sd=1) $ \\
Noise 3: $\epsilon_3 \sim N(mean=1, sd=5)$, \\
and the noise coefficient $\tau \in \{0.25,0.5,1,2\}$.
\\
We compare \collapsingflame\ to FLAME. 

\begin{figure*}[t]
\centering
\includegraphics[width=\linewidth]{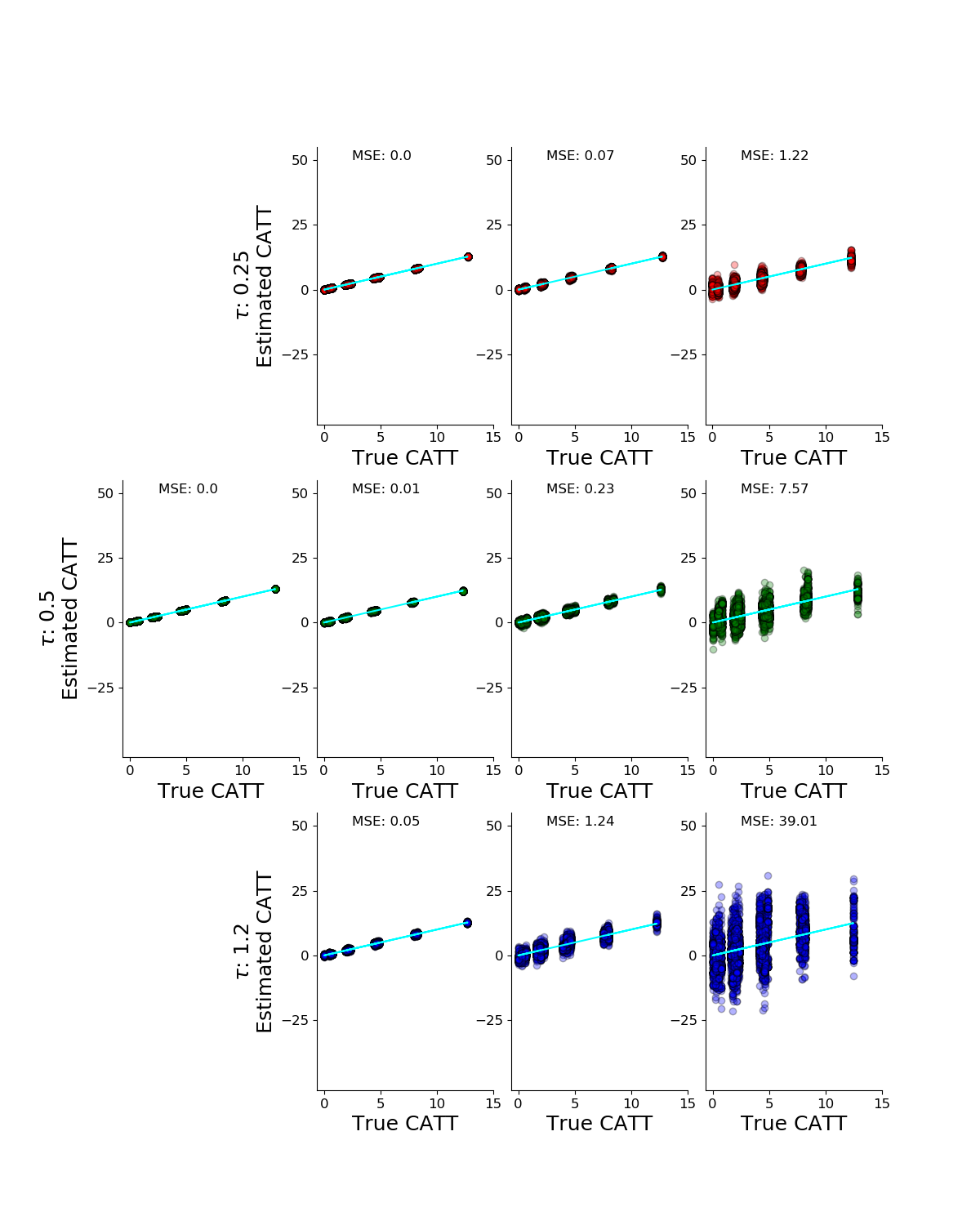}
\vspace{.3in}
\caption{Results of \collapsingflame\ for different noise levels. Each row represents a different $\tau$ and each column represents a different $\epsilon$\label{fig:dame_noise}}
\end{figure*}

\begin{figure*}[t]
\centering
\includegraphics[width=\linewidth]{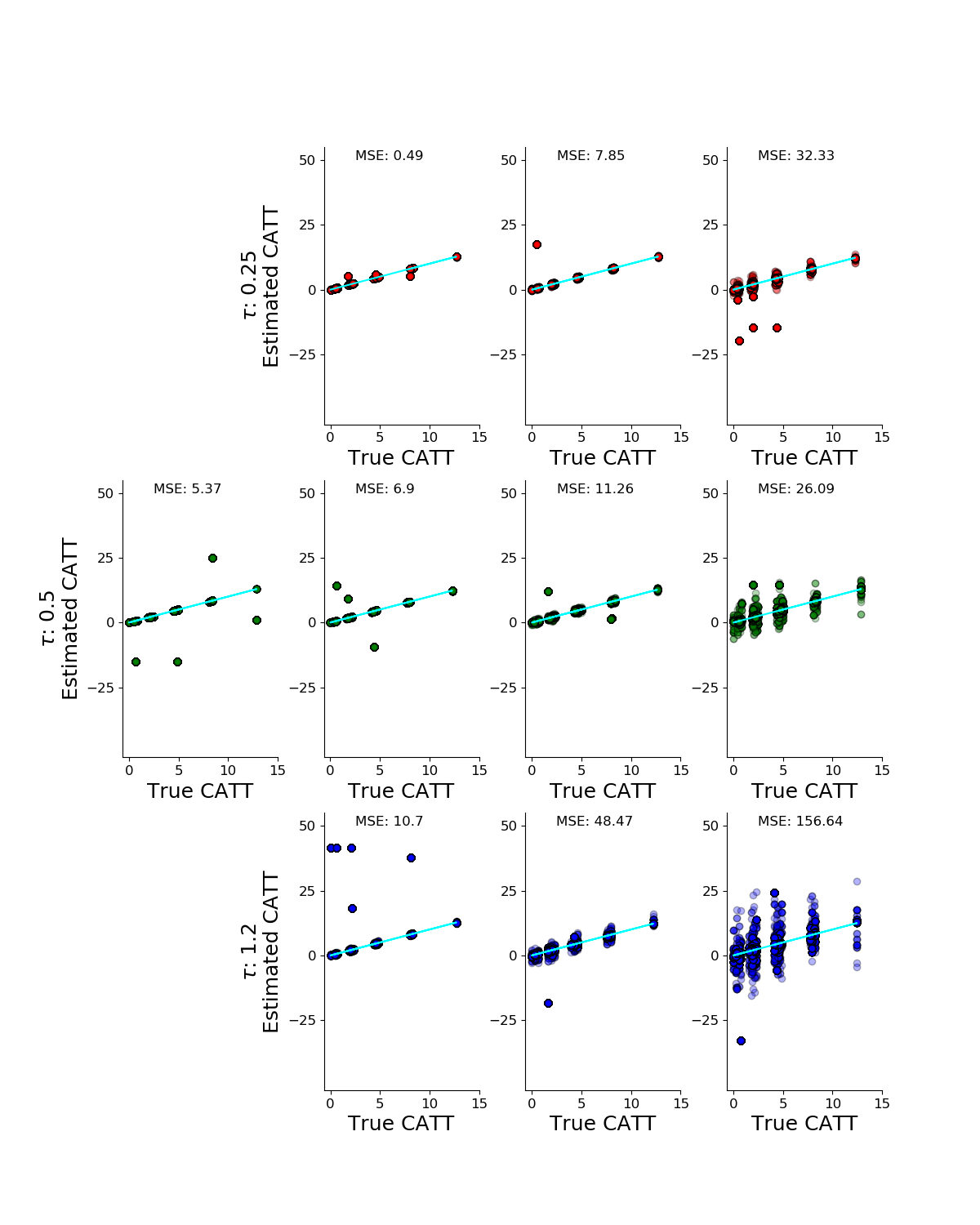}
\vspace{.3in}
\caption{Results of FLAME for different noise levels. Each row represents a different $\tau$ and each column represents a different $\epsilon$\label{fig:flame_noise}}
\end{figure*}

{\bf Results:} As Figures~\ref{fig:dame_noise} and \ref{fig:flame_noise} show,  \collapsingflame\ outperforms FLAME at all noise levels. While both methods degrade in quality (except for $\tau = 0.5$) as the amount of noise increases, the degradation in \collapsingflame\ is smaller than that of FLAME.  
Further smoothing of the output of \collapsingflame\ would likely lead to a further reduction in the influence of measurement noise. 
\section{Details of Breaking the Cycle of Drugs and Crime Study}\label{sec:btc-details}
\subsection{Details About Survey}
A survey was conducted in Alabama, Florida, and Washington regarding the program's effectiveness, with high quality data for over 380 individuals. These data (and this type of data generally) can be a powerful tool in the war against opioids, and our ability to draw interpretable, trustworthy conclusions from it depends on our ability to construct high-quality matches. For the survey, participants were chosen to receive screening shortly after arrest and participate in a drug intervention under supervision. Similar defendants before the start of the BTC program were selected as the control group. Features are listed in Table \ref{table:btcfeatures}.

\begin{table*}[ht]
\begin{center}
\caption{\small Features for BTC data.\label{table:btcfeatures}}
\begin{tabular}{|p{9cm}|} 
\hline
Feature  \\ 
\hline
1. Live with anyone with an alcohol problem   \\
\hline
2. Have trouble understanding in life  \\
\hline 
3. Live with anyone using non prescription drugs  \\
\hline
4. Have problem getting along with father in life  \\
\hline
5. Have an automobile  \\
\hline
6. Have drivers license  \\ 
\hline
7. Have serious depression or anxiety in past 30 days  \\
\hline
8. Have serious anxiety in life  \\
\hline
9. SSI benefit last 6 months \\
\hline
10. Have serious depression in life  \\
\hline
\end{tabular}
\end{center}
\end{table*}


\begin{table*}[ht]
\begin{center}
\caption{\small Order in which features were processed for \collapsingflame\ and FLAME. The feature numbers correspond to the feature numbers in Table~\ref{table:btcfeatures}. The number in the parenthesis corresponds to the number of units matched for the first time at that round. Before any covariates are dropped, 287 individuals are matched on all features, which is 75\% of the data. \label{table:featureorder}}
\begin{tabular}{|c|l|l|} 
\hline
& \collapsingflame  & FLAME   \\ [0.5ex]
\hline
1st & 4: problem with father  {(15 new units matched)} & 4 (7 units)\\
\hline
2nd & 5: have an automobile (9 units)& 4,7 (25 units)\\
\hline
3rd & 7: have serious depression (24 units)& 4,7,9 (9 units)\\
\hline
4th & 4,7  (3 units)& 4,7,9,1 (7 units)\\
\hline
5th & 5,7 (1 unit) & 4,7,9,1,8 (12 units)\\
\hline
6th & 4,5 (7 units) & 4,7,9,1,8,10 (6 units)\\
\hline
7th & 4,5,7  (0 units) &  4,7,9,1,8,10,6 (5 units)\\
\hline
8th & 9 (8 units) & 4,7,9,1,8,10,6,5 (11 units) \\
\hline
9th & 4,9 (0 units)& 4,7,9,1,8,10,6,5,2 (5 units)\\
\hline
$\vdots$ & & \\
\hline
196th & 1,2,4,5 (1 unit) &\\
\hline
\end{tabular}
\end{center}
\end{table*}


\subsection{Order of Dropping Covariates}
For both \collapsingflame\ and FLAME we used ridge regression as the machine learning method for the Full-AME problem, calculating variable importance as the difference in mean squared error before and after dropping the variable.
The order in which \collapsingflame\ and FLAME process covariates could be different. Table \ref{table:featureorder} shows the order in which the dynamic versions of the two algorithms process the covariates. 
The first covariate that the two algorithms process is identical: ``Have problem getting along with father in life'' but the two diverge afterwards. 
At the second round, \collapsingflame\ processes the covariate ``Have an automobile.'' 
On the other hand, at that same second round, FLAME processes ``Have serious depression or anxiety in past 30 days'', which now is dropped along with ``Have problem getting along with father in life.'' What is important is that \collapsingflame\ is able to construct matched groups by only dropping subsets of what FLAME drops as early as the second and third iteration of the algorithm.

\begin{figure}[h]
\centering
\includegraphics[width=.45\textwidth]{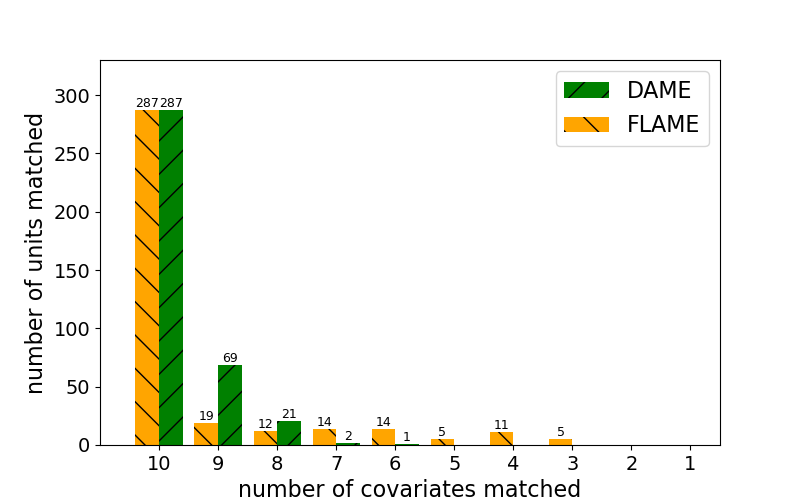}
\caption{\small Number Matched: Number of units matched per covariates for the BTC data
\label{fig:btcnummatched}}
\end{figure}
\subsection{Match Quality for FLAME and \collapsingflame}
We compare the quality of matches in the BTC data between FLAME and \collapsingflame\ in terms of the number of covariates used to match within the groups. Many of the units matched exactly on all covariates and thus were matched by both algorithms at the first round. In fact 75\% of the data are matched on all covariates. This is important, because exact matching alone yields the highest quality CATE estimates for most of the data; if we had used a classical propensity score matching technique, we may not have noticed this important aspect of the data.

For the remaining units that do not have exact matches on all covariates, \collapsingflame\ matches on more covariates than FLAME. In Figure \ref{fig:btcnummatched} we see that \collapsingflame\ matched many more units on 9 out of the 10 variables than FLAME; FLAME cannot match the same data on so many variables. 


\subsection{CATEs from BTC analysis}
\begin{figure}[]
\centering
\includegraphics[width=.48\textwidth]{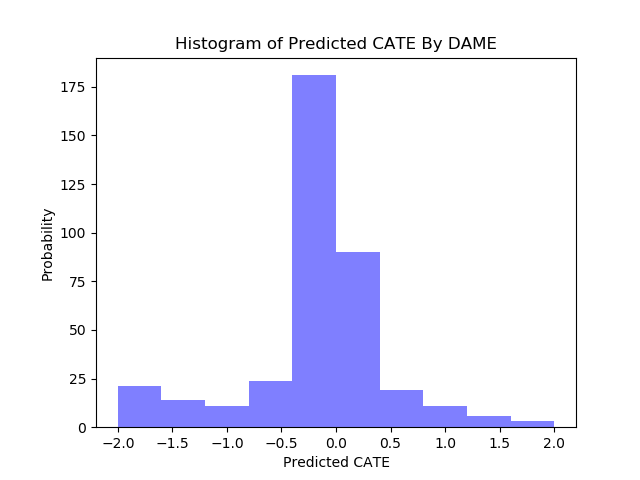}
\caption{\small Histogram of estimated CATE by \collapsingflame. For individuals where the CATE is negative, it means that BTC was estimated to reduce crime. 
\label{fig:catesforBTC}}
\end{figure}

We plot a histogram of the estimated CATEs for BTC in Figure \ref{fig:catesforBTC}. The program does not seem to provide uniform protection from future arrests, but does seem to protect some individuals. The majority of people are estimated to experience little to no effect from the program. 

\subsection{A Comparison of \collapsingflame\ with SVM-Based Method Minimax Surrogate Loss}\label{sec:dameVSsvm}
\begin{figure}[]
\centering
\includegraphics[width=.45\textwidth]{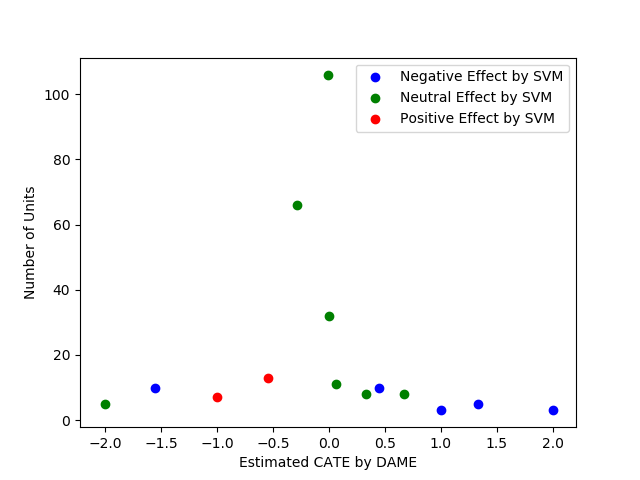}
\vspace{.3in}
\caption{Comparison between \collapsingflame\ and SVM-based method
     \label{fig:ccb}}
\end{figure}

We can use \collapsingflame\ as a tool to check the performance of a black box machine learning approach. 
We chose a recent method that predicts whether treatment effects are positive, negative, or neutral, using a support vector machine formulation \citep{minsurrogateloss2018}. 
We ran \collapsingflame\ on the BTC dataset and saved the CATE for each treatment and control unit that were matched. Units with a positive CATE (outcome on treatment unit minus outcome on control unit) are considered to have a negative treatment effect, meaning that the program increased the probability of crime. Units with a negative CATE analogously had a positive predicted treatment effect. We also implemented the SVM approach and recorded a prediction of positive, negative, or neutral treatment effect for each unit. 
Figure~\ref{fig:ccb} plots the CATEs for all the units that were matched exactly by \collapsingflame\ and colors them according to the output of the SVM. 
Since the distribution of some covariates is unbalanced, the number of matched groups is small with most units belonging to large groups.

Figure \ref{fig:ccb} shows that \collapsingflame\ and the SVM approach agree on the direction of the treatment effect for most of the matched units:
Most positive CATEs corresponded to negative treatment effects from the SVM. Only two points have a mismatch between \collapsingflame\  and SVM: the left-most green (neutral) labeled and blue (negative) labeled points. 

The easiest way to explain the discrepancy between the two methods is that \collapsingflame\ is a matching method, not a statistical model and so does not smooth CATEs. CATEs are sometimes computed using a very small number of units, so it is possible that the SVM simply smoothed out the treatment effect estimates so that there was a different predicted treatment effect on some of the units.
\cut{\begin{table*}[ht]
\begin{center}
\caption{\small Correlation between each feature and predicted outcomes by \collapsingflame\ . \label{table:featureorder}}
\begin{tabular}{|p{1.5cm}|p{3cm}|p{3cm}|} 
\hline
Features & Positive Predicted CATEs & Negative Predicted CATEs  \\ [0.5ex]
\hline
1st & -0.11 & 0.02\\
\hline
2nd & -0.24 &  0.14\\
\hline
3rd & -0.12 & 0.13 \\
\hline
4th & 0.05 & -0.09 \\
\hline
5th & 0.13 & -0.13 \\
\hline
6th & 0.17 & -0.47 \\
\hline
7th & 0.28 & -0.29 \\
\hline
8th & 0.36 & -0.41 \\
\hline
9th & 0.11 & -0.31 \\
\hline
10th & 0.04 & 0.01\\
\hline
\end{tabular}
\end{center}
\end{table*}}
To evaluate this hypothesis,
we computed the Hamming distance between the special group's units (this is the group where \collapsingflame\ and the SVM disagree) with units in other groups to investigate.

In Figure \ref{fig:ccb}, the units within the leftmost blue (negative) labeled matched group were much closer to other blue (negative) labeled matched groups than to green (neutral) or red (positive) labeled groups, suggesting that smoothing the estimates after running \collapsingflame\ would likely make them consistent with the SVM results.
The units within the leftmost green (neutral) labeled matched group are not closer to other green (neutral) labeled matched groups than other colors, suggesting that neither SVM nor \collapsingflame\ have information to properly identify the causal effect for this group. 
We similarly investigated the blue (negative) labeled group for which CATE$=0.5$ and again, the covariate values of its units were closer in Hamming distance to other blue (negative) labeled groups than to other points. Thus, additional smoothing of the CATEs from the matched groups could likely yield estimated positive and negative treatment effects similar to those of the SVMs.

\eat{

{\color{red} {\LARGE COMMENT OUT ALL\\

FOLLOWING SECTIONS\\

IN THE MAIN BODY}}

\section{Adjustment of weights}\label{sec:adjust-weights}
\textcolor{red}{Cynthia thinks we should remove this section after we add its main components back into Sec 4}

One noticeable feature of  \collapsingflame\  resides in its ability to dynamically adjust the weight of the covariates throughout the run of the algorithm. This property is important as covariates become more or less relevant in presence of other covariates. To recompute the weight ie the importance score of each covariate,  \collapsingflame\ utilizes a hold-out training set $D^H$ whose covariate space is restricted to the subset of the covariates being considered ($\mathcal{J} \setminus s$).  Using the hold-out set $D^H=[X^H, Y^H, T^H]$ that includes both treatment and control units, we obtain the vectors representing the weights $\beta_t$ and $\beta_c$ by linear regressions on control and treatment units. The readjusted weights are used to compute a {\it prediction error $\mathcal{PE}$} measure to determine which subset of covariates leads to the best estimation quality. Specifically, $\mathcal{PE}$ is computed for each active set considered by
\begin{equation}\label{equn:PE}
\begin{split}
\PE(D^H_{\Theta,s}) = \qquad\qquad\qquad\qquad\qquad\qquad\qquad\qquad\qquad &\\ 
\min_{\beta_t,\beta_c}\left[ 
\frac{1}{\sum (1-T^{H}_u)}\sum_{u:T_u^{H}=0} \left(Y_u^H-X^H\left(u, J \ s \right)\beta_c\right)^2\ \right.&\\
\left.+\frac{1}{\sum T^{H}_u}\sum_{u:T_u^{H}=1}\left(Y_u^H-X^H\left(u, J \ s \right)\beta_t\right)^2 \right] 
\end{split}
\end{equation}

{\bf Notation}: $\mathbf{D}$ is the input dataset, $\mathbf{D^H}$ is the holdout dataset, $\Theta$ represents a selector function on the rows and columns of the input dataset for any given matched unit $u$. In other words, $D_{\Theta,u}$ consists of a subset of rows and covariates that match with $u$ (maximizing $\Theta$ gives the main matched group for any unit $u$), $T^{H}_u$ and $T^{H}_u$ are respectively the treatment indicator and the outcome for unit $u$  in the hold-out dataset.

In addition to optimizing over $\vv^T \mathbf{w}$, we also give preference to subsets of attributes that allow better balancing of treatment and control units in the matched group. That balancing factor $\mathcal{BF}$ depends only on the number of units matched at each iteration by dropping $s$ and the remaining unmatched units. 
\begin{eqnarray}
\mathcal{BF}_{\Theta,s} = \frac{ \#  \textrm{\rm of matched controls by dropping } s}{ \#  \textrm{\rm  of remaining controls}} + \nonumber\\
\frac{ \#  \textrm{\rm of matched treatments by dropping } s}{ \#  \textrm{\rm  of remainging treatments}}
\end{eqnarray}

A matched quality quantity consisting of a trade off between these two measures ensures that \collapsingflame\ only processes sets that maintain good prediction quality while the matched groups contain enough units to yield good estimate of treatment effects.

{\color{red}{\LARGE DO NOT COMMENT OUT BTC DETAILS}}
}

\end{document}